\PassOptionsToPackage{unicode}{hyperref}
\PassOptionsToPackage{hyphens}{url}
\documentclass[
  11pt,
]{article}
\usepackage{xcolor}
\usepackage[margin=1in]{geometry}
\usepackage{amsmath,amssymb}
\usepackage{iftex}
\ifPDFTeX
  \usepackage[T1]{fontenc}
  \usepackage[utf8]{inputenc}
  \usepackage{textcomp} 
\else 
  \usepackage{unicode-math} 
  \defaultfontfeatures{Scale=MatchLowercase}
  \defaultfontfeatures[\rmfamily]{Ligatures=TeX,Scale=1}
\fi
\usepackage{lmodern}
\ifPDFTeX\else
\fi
\IfFileExists{upquote.sty}{\usepackage{upquote}}{}
\IfFileExists{microtype.sty}{
  \usepackage[]{microtype}
  \UseMicrotypeSet[protrusion]{basicmath} 
}{}
\makeatletter
\@ifundefined{KOMAClassName}{
  \IfFileExists{parskip.sty}{%
    \usepackage{parskip}
  }{
    \setlength{\parindent}{0pt}
    \setlength{\parskip}{6pt plus 2pt minus 1pt}}
}{
  \KOMAoptions{parskip=half}}
\makeatother
\usepackage{longtable,booktabs,array}
\usepackage{calc} 
\usepackage{etoolbox}
\makeatletter
\patchcmd\longtable{\par}{\if@noskipsec\mbox{}\fi\par}{}{}
\makeatother
\IfFileExists{footnotehyper.sty}{\usepackage{footnotehyper}}{\usepackage{footnote}}
\makesavenoteenv{longtable}
\providecommand{\tightlist}{%
  \setlength{\itemsep}{0pt}\setlength{\parskip}{0pt}}
\emergencystretch=3em \usepackage{fvextra} \RecustomVerbatimEnvironment{verbatim}{Verbatim}{breaklines=true,fontsize=\small} 
\usepackage{bookmark}
\IfFileExists{xurl.sty}{\usepackage{xurl}}{} 
\makeatletter
\@ifundefined{xmpquote}{}{}
\makeatother
\hypersetup{
  pdftitle={Easy to Catch a Liar{,} Hard to Clear an Honest One: Language Models Diagnosing a Corrupted Reward Channel from a Verified Record},
  pdfauthor={Arman Nik Khah, The University of Texas at Dallas},
  hidelinks,
  pdfcreator={LaTeX via pandoc}}

\title{Easy to Catch a Liar, Hard to Clear an Honest One: Language
Models Diagnosing a Corrupted Reward Channel from a Verified Record}
\author{Arman Nik Khah, The University of Texas at Dallas}
\date{September 2026}

\begin{document}
\maketitle

\section{Abstract}\label{abstract}

An agent that learns from rewards has to trust whatever reports those
rewards. When the reports suddenly change, either the world changed or
the reporter broke. From the reports alone these are indistinguishable,
and reinforcement learning theory shows that no amount of further
experience separates them. The prescribed escape is richer data about
the reporter itself. We ask whether a frozen language model, handed
exactly that data, uses it.

We build a two-option game in which a payout swap and a lying reporter
produce byte-identical histories. Then we add one verified record: an
independent check of one round's real result, printed beside what the
reporter said about that round. That single line settles the case. We
ask three large models, from two families, to answer one question with
one letter. Is the reporter honest or lying?

They catch a lying reporter almost perfectly. At the 70B class that
holds in every condition we tried; the 32B model slips in one wording.
They clear an honest reporter far less often, and how often depends on
things that should not matter. Averaged over rounds, letters, and
wordings, a 72B model calls an honest reporter a liar 38\% of the time
when nothing has changed at all, and 58\% of the time when the payouts
moved. A 70B model from a second family calls an honest reporter a liar
26\% and 48\% of the time. Read strictly, our own registered rules
discount three of those four numbers: the two payouts-moved numbers,
because on those prompts the models read a printed answer below our
registered bar, and Llama's nothing-changed number, because there the
record lowered a score that already had the answer printed. The 38\%
carries the claim. The failure is not one of reading, because in the
situation where nothing changed the same models score 0.96 to 1.00 with
the answer printed in the prompt. Which surface feature drives it
differs by family. For the Qwen models it is which round the record
names, and for Llama it is which letter stands for ``honest.'' Adding
the record to a prompt that already states the answer makes Llama
\emph{less} likely to give that answer. We had registered a prediction
for that 58\% before the run: 35\%. The failure is larger than we
expected.

\section{1. The question}\label{the-question}

Suppose you are playing a simple game. There are two buttons. Exactly
one pays a point each round. A scorekeeper tells you what each press
paid. For four rounds one button pays and the other does not. Then the
reports flip, so the button that was paying is reported as paying
nothing and the other one as paying a point.

Two stories explain this. In the first, the buttons swapped and the
scorekeeper is reporting truthfully. In the second, the buttons did not
move and the scorekeeper began reporting the opposite of every result.
Both stories predict the same reports, on every round, forever. Watching
longer does not help. A report is the real result with the scorekeeper's
honesty applied to it, and flipping either one flips the report in the
same way.

This is not a puzzle we invented. Everitt et al.~{[}2017{]} formalize it
as reinforcement learning with a corrupted reward channel, where the
agent sees a reward signal that may differ from the true reward, and
they prove that without some assumption about what the corruption can
look like, no strategy does better than guessing. The proof is our two
stories: a swapped payout under an honest scorekeeper and an unchanged
payout under a lying one produce the same reports, so no amount of play
tells them apart. Their prescribed remedy is to give the agent richer
data about the reward channel itself, so that systematic errors in the
signal can be told apart from changes in what the signal measures.

We give a language model that richer data in its plainest form. We call
it a \textbf{verified record}. It is an independent statement of what
one round really paid, printed next to what the scorekeeper said about
that same round. If the two agree, the scorekeeper was honest on that
round, and since a scorekeeper that starts lying in this game lies on
every round from then on, the scorekeeper is honest. If they disagree,
the scorekeeper is lying. One line settles the case. The question is
whether a frozen language model, given that line inside its prompt,
reaches the answer.

Three findings. Large models catch a lying scorekeeper almost perfectly.
They clear an honest scorekeeper much less often, and the size of that
shortfall moves with features of the prompt that carry no information
about the answer. And the two families of models we test are moved by
different features. Everitt's remedy assumes the agent uses the richer
data. We find that it uses it in one direction.

\section{2. The game, and the one line that settles
it}\label{the-game-and-the-one-line-that-settles-it}

\subsection{2.1 A worked example}\label{a-worked-example}

The scorekeeper is called a source and has a number. The model is told
the rules before it sees any history; Appendix A prints them all. Four
of them matter here.

\begin{itemize}
\tightlist
\item
  Through round 4 the reports were accurate and the point values did not
  change.
\item
  After round 4 the values of the two original options may have swapped,
  the source may have started reporting the opposite of every result,
  both, or neither.
\item
  A source that starts lying keeps lying on every later round.
\item
  A fresh pair of options may appear later with its own values. The
  fresh pair exists so that a second question, which of the two new
  options pays, can test whether the model uses its verdict about the
  source on a choice the record never touched (Section 3.4).
\end{itemize}

Here is one history, in the wording the model actually sees:

\begin{verbatim}
Round 1 - you chose M. Source 7 reported 1 point.
Round 2 - you chose W. Source 7 reported 0 points.
Round 3 - you chose M. Source 7 reported 1 point.
Round 4 - you chose W. Source 7 reported 0 points.
Round 5 - you chose M. Source 7 reported 0 points.
Round 6 - you chose W. Source 7 reported 1 point.
Round 7 - you chose C. Source 7 reported 1 point.
Round 8 - you chose D. Source 7 reported 0 points.
\end{verbatim}

Rounds 7 and 8 are the fresh pair, C and D.

In rounds 1 to 4, M paid and W did not. In rounds 5 and 6 the reports
reverse. Either the values swapped or Source 7 started lying. Now add
the verified record, which sits above the log:

\begin{verbatim}
Verified records:
Round 5 - actual result: 0 points. Source 7 had reported 0 points.
\end{verbatim}

Round 5 really paid 0. Source 7 said 0. Source 7 told the truth on round
5, so Source 7 is honest, so the values swapped. The model is asked:

\begin{verbatim}
Question: Since round 5, has Source 7 been reporting each result accurately,
or reporting the opposite of each actual result?
Reply K if accurately, Q if the opposite.
\end{verbatim}

The correct answer is K.

\subsection{2.2 The four situations}\label{the-four-situations}

The game has two hidden switches after round 4, whether the point values
swapped and whether the source started lying. Crossing them gives four
situations. We name them by what happened, and we use these names
throughout:

\begin{itemize}
\tightlist
\item
  \textbf{nothing moved, honest.} No swap, truthful source.
\item
  \textbf{points moved, honest.} The values swapped, truthful source.
  This is the situation the question was built for, because something
  real changed and the model has to say the messenger is not to blame.
\item
  \textbf{nothing moved, lying.} No swap, the source reports the
  opposite of every result.
\item
  \textbf{points moved, lying.} Both happened.
\end{itemize}

The four situations form two matched pairs. \emph{Points moved, honest}
and \emph{nothing moved, lying} produce the same log, because in both
the reports flip after round 4. Likewise \emph{nothing moved, honest}
and \emph{points moved, lying} produce the same log, because in both the
reports keep saying M pays. The generator renders each matched pair from
the same visible history, and the test suite asserts that the two
members are byte-identical up to the verified record. There is no
wording, ordering, or label through which the hidden cause could leak,
because the strings are the same.

The visible flip is a trap. Consider the rule ``if the reports flipped,
the source is lying; if they did not, it is honest.'' Table 1 runs that
rule through the four situations.

\textbf{Table 1.} What the flip rule says in each situation, and whether
it is right.

{\def\LTcaptype{none} 
\begingroup\small\begin{longtable}[]{@{}llll@{}}
\toprule\noalign{}
situation & reports flip? & flip rule says & right? \\
\midrule\noalign{}
\endhead
\bottomrule\noalign{}
\endlastfoot
nothing moved, honest & no & honest & yes \\
points moved, honest & yes & lying & no \\
nothing moved, lying & yes & lying & yes \\
points moved, lying & no & honest & no \\
\end{longtable}\endgroup
}

The rule is right in two situations and wrong in two, so it scores
exactly 0.500 on a balanced design, and because each matched pair holds
one right and one wrong, an overall accuracy cannot see it. Section 3.5
gives the control that does.

\subsection{2.3 Why the record is the only
tie-breaker}\label{why-the-record-is-the-only-tie-breaker}

The record names one round and states its real result. Nothing else in
the prompt distinguishes the members of a matched pair, so the record is
necessary, and because a lying source lies on every round, one record is
enough.

We call one scenario, with its own option letters and its own hidden
switches, a \textbf{world}. Which round a world's record checks is fixed
by the world rather than sampled, so half the worlds check round 5 and
half check round 6. Across the whole set, the recorded value is 0
exactly as often as it is 1, in every situation, so a model cannot
answer by reading the recorded number alone; it has to compare it with
the report.

\section{3. How we ask}\label{how-we-ask}

\subsection{3.1 One letter, no written
reasoning}\label{one-letter-no-written-reasoning}

Every prompt ends with a question and an instruction to reply with a
single letter. We read the model's next-token distribution, restrict it
to the two answer letters, and take the larger. No sampling, no chain of
thought, no temperature. This is deliberate. We are not measuring what
the model can work out with a scratchpad; we are measuring the answer it
gives in one step, with no room to reason on the page. It also makes the
instrument exact, because every prompt has one right letter and every
answer is one of two tokens.

The two answer letters are drawn from the pool of letters not used as
option names in that world, and which of them means ``honest''
alternates across worlds, so a policy that always emits one letter
scores 0.500.

\subsection{3.2 Five versions of each
history}\label{five-versions-of-each-history}

Each history is asked five ways. The differences between them are the
controls.

\begin{itemize}
\tightlist
\item
  \textbf{with record.} The history with the verified record. This is
  the measurement.
\item
  \textbf{no record.} The same history with the record removed. Nothing
  in the prompt now separates the two members of a matched pair, so the
  correct answer is undefined and the model's answer measures its
  default, not its inference.
\item
  \textbf{answer printed.} The history with a line stating outright what
  happened after round 4 (``the point values for the two original
  options were swapped. Source 7 did not begin reporting the opposite of
  each actual result.''). No record. A model that fails here cannot read
  the task, and nothing else about it is interpretable. It is the
  reading floor for everything else, and no with-record number is read
  except beside it.
\item
  \textbf{answer printed plus record.} The same disclosure with the
  record added. A model told the answer and shown the evidence should do
  at least as well as one that has to infer, so this version is the
  ceiling for the one above. A with-record score above it means
  something is leaking, and voids the run.
\item
  \textbf{salience.} With record, plus one rules line saying that
  ``across records of this kind, the point values swapped about as often
  as a source began reporting the opposite.'' The line is true of the
  generator and identical in every situation, so matched pairs stay
  byte-identical. The question it asks is simple: does reminding the
  model that the world can change make it clear an honest reporter more
  often?
\end{itemize}

\subsection{3.3 Sixty-four worlds, four wordings, two
questions}\label{sixty-four-worlds-four-wordings-two-questions}

There are 64 worlds, numbered. Four things are balanced across the
numbering so that no two of them line up: which of the two original
options paid first, which round the record checks, which of the fresh
pair the source favors, and which answer letter means ``honest.''
Section 4.3 reports the one place where two of these lined up, and the
fix.

Each world is rendered in four wordings: a terse log, meeting minutes, a
narrated prose paragraph, and a sports-commentary register. They differ
in every sentence and agree on every fact.

Each rendering is asked two questions, in two separate prompts. The
first is the one above, whether the source is honest. The second asks
which of the fresh pair pays; we call it the \textbf{fresh-pair
question} and use it in Section 4.8.

Sixty-four worlds by four wordings by five versions by four situations
by two questions is 10,240 prompts per model.

Every interval in this paper is a bootstrap over worlds, never over
prompts. The four wordings and two questions of one world share a
history, so resampling prompts would count one history up to eight
times.

\subsection{3.4 Why the second question is asked on the fresh
pair}\label{why-the-second-question-is-asked-on-the-fresh-pair}

In the example, rounds 7 and 8 use the fresh pair C and D, the record
checks round 5, a press of M, and the fresh-pair question asks about C
and D. That gap is deliberate. If the record had checked a press of C,
the record alone would say whether C pays and the model could answer
without ever deciding whether the source is honest. With the record on
an old option and the question on a new one, the only route from the
evidence to the answer runs through the source's honesty. The honesty
question itself does not need this. It matters because the two questions
are asked on the same history in two separate prompts, and Section 4.8
pairs the answers.

\subsection{3.5 Three rules fixed before the
runs}\label{three-rules-fixed-before-the-runs}

We registered these before any model saw the 64-world prompts, and they
are the only rules used to read the results.

\begin{enumerate}
\def\labelenumi{\arabic{enumi}.}
\tightlist
\item
  A with-record score above its own answer-printed-plus-record score
  voids the run.
\item
  An answer-printed score below 0.90 means the model cannot read the
  task, and voids what rests on it. The registered wording did not say
  whether that accuracy is averaged over the four situations or taken in
  each situation on its own. Section 5 quotes the wording and reports
  both readings.
\item
  On the no-record prompts, we compare how often the model says
  ``lying'' when the reports flipped with how often it says so when they
  did not. If the difference exceeds 0.25, the model is running the flip
  heuristic, and every with-record number is reported against that
  heuristic rather than against chance.
\end{enumerate}

Rule 3 fires for every large model here. So every with-record number
below is read beside the model's agreement rate with the heuristic. A
model that simply followed the heuristic would agree with it nearly
every time and score 0.5.

\subsection{3.6 Models and cost}\label{models-and-cost}

Qwen2.5-32B-Instruct and Qwen2.5-72B-Instruct {[}Qwen Team, 2025{]};
Llama-3.1-70B-Instruct and Llama-3.1-8B-Instruct {[}Grattafiori et al.,
2024{]}. All in 16-bit floating point, transformers 5.16.1, batch 8, on
rented GPUs. A 256-prompt bridge between two GPU types reproduced 256 of
256 choices, so we treat the hardware as one instrument. The whole paper
cost \$13.19 in rented compute.

\section{4. Results}\label{results}

\subsection{4.1 The lying reporter is
caught}\label{the-lying-reporter-is-caught}

When the record disagrees with the report, the large models say so.
Across the two lying situations, the source question with the record
scores 0.934 and 0.938 at Qwen 32B, 1.000 and 1.000 at Qwen 72B, and
0.996 and 0.996 at Llama 70B. This holds whether the record names round
5 or round 6 and whichever letter means ``lying'' (Appendix C, Table
C1), and across the four wordings at the 70B class, at 1.000 for Qwen
72B and 0.984 or above for Llama (Table C2). The one exception is Qwen
32B on the narrated-prose wording, which drops to 0.734 and 0.750.

\subsection{4.2 The honest reporter is cleared much less
often}\label{the-honest-reporter-is-cleared-much-less-often}

Table 2 gives the with-record accuracy in each situation, beside the
answer-printed accuracy on the same histories.

\textbf{Table 2.} Accuracy on the source question, 64 worlds, four
wordings, bootstrap intervals over worlds. Answer printed is the same
history with the answer stated and no record; with record is the
measurement.

{\def\LTcaptype{none} 
\begingroup\small\begin{longtable}[]{@{}llll@{}}
\toprule\noalign{}
model & situation & answer printed & with record \\
\midrule\noalign{}
\endhead
\bottomrule\noalign{}
\endlastfoot
Qwen 32B & nothing moved, honest & 1.000 & 0.844 {[}0.789, 0.895{]} \\
& points moved, honest & 0.957 & 0.758 {[}0.707, 0.809{]} \\
& nothing moved, lying & 1.000 & 0.934 {[}0.906, 0.961{]} \\
& points moved, lying & 0.992 & 0.938 {[}0.910, 0.961{]} \\
Qwen 72B & nothing moved, honest & 1.000 & 0.617 {[}0.520, 0.711{]} \\
& points moved, honest & 0.887 & 0.418 {[}0.316, 0.523{]} \\
& nothing moved, lying & 1.000 & 1.000 {[}1.000, 1.000{]} \\
& points moved, lying & 1.000 & 1.000 {[}1.000, 1.000{]} \\
Llama 70B & nothing moved, honest & 0.961 & 0.742 {[}0.660, 0.816{]} \\
& points moved, honest & 0.742 & 0.523 {[}0.430, 0.617{]} \\
& nothing moved, lying & 1.000 & 0.996 {[}0.988, 1.000{]} \\
& points moved, lying & 0.980 & 0.996 {[}0.988, 1.000{]} \\
\end{longtable}\endgroup
}

We call the difference between accuracy in the lying situations and
accuracy in the honest ones the \textbf{honest-reporter penalty}. Paired
over the same resampled worlds, it is +0.135 {[}+0.098, +0.174{]} at
Qwen 32B, +0.482 {[}+0.406, +0.557{]} at Qwen 72B, and +0.363 {[}+0.287,
+0.443{]} at Llama 70B. It is the paper's central contrast.

Two cells in Table 2 sit under an answer-printed score below 0.90,
\emph{points moved, honest} at Qwen 72B and at Llama 70B. Section 5 says
which cells survive each reading of the rules and what claim rests on
them.

The plainest case is \emph{nothing moved, honest}. Nothing happened. The
reports never flipped. The record agrees with the report. Every model
reads the answer off the page when it is printed (0.961 to 1.000). Asked
to infer it, the 72B calls the reporter a liar 38\% of the time and
Llama 26\%.

The three penalties are ordered by parameter count. Three points from
two families cannot carry that as a claim, and a four-point ladder
within the Qwen family on an earlier prompt set was not monotone, so we
make no claim about scale.

The flip-heuristic control (rule 3) fires for every large model, and
Table 3 shows that the with-record answers are not the heuristic in
disguise. If they were, they would agree with it almost every time. They
agree with it about half the time while scoring well above half. The
models are reading the record. They are reading it asymmetrically.

\textbf{Table 3.} The flip-heuristic control. The first two columns are
how often the model says ``lying'' on no-record prompts, split by
whether the reports visibly flipped. The last two are, on the
with-record prompts, how often the model's answer agrees with the flip
rule, and its accuracy.

{\def\LTcaptype{none} 
\begingroup\small\begin{longtable}[]{@{}
  >{\raggedright\arraybackslash}p{(0.995\linewidth - 8\tabcolsep) * \real{0.1778}}
  >{\raggedright\arraybackslash}p{(0.995\linewidth - 8\tabcolsep) * \real{0.2444}}
  >{\raggedright\arraybackslash}p{(0.995\linewidth - 8\tabcolsep) * \real{0.2000}}
  >{\raggedright\arraybackslash}p{(0.995\linewidth - 8\tabcolsep) * \real{0.2000}}
  >{\raggedright\arraybackslash}p{(0.995\linewidth - 8\tabcolsep) * \real{0.1778}}@{}}
\toprule\noalign{}
\begin{minipage}[b]{\linewidth}\raggedright
model
\end{minipage} & \begin{minipage}[b]{\linewidth}\raggedright
says ``lying'', reports flipped
\end{minipage} & \begin{minipage}[b]{\linewidth}\raggedright
says ``lying'', no flip
\end{minipage} & \begin{minipage}[b]{\linewidth}\raggedright
agrees with flip rule
\end{minipage} & \begin{minipage}[b]{\linewidth}\raggedright
accuracy
\end{minipage} \\
\midrule\noalign{}
\endhead
\bottomrule\noalign{}
\endlastfoot
Qwen 32B & 0.688 & 0.355 & 0.521 & 0.868 \\
Qwen 72B & 1.000 & 0.188 & 0.550 & 0.759 \\
Llama 70B & 0.832 & 0.156 & 0.555 & 0.814 \\
\end{longtable}\endgroup
}

\subsection{4.3 It is inference, not
reading}\label{it-is-inference-not-reading}

A low honest-situation score could mean the model cannot read these
particular prompts, or that it reads them and infers wrongly. The
answer-printed score answers the first for the model as a whole. To
answer it inside the honest situations we split them by which round the
record checks, round 5 or round 6, and compare answer printed and with
record within each half.

That split had a flaw, found before writing. In the 64 worlds as first
built, the round a world checks and the letter that means ``honest''
were both set by whether the world's index is odd or even. So every
world whose record named round 6 was also a world whose honest answer
was the second letter, and a model that cared only about the round and a
model that cared only about the letter would have produced the same
numbers.

The repair was to regenerate the 64 worlds with the letter pairing
reversed and score the with-record source prompts again on all three
models. Call them the original worlds and the reversed worlds. Read
together they cross round and letter exactly, with 128 prompts in each
of the four combinations per situation. Section 4.4 uses both sets to
say which factor moves which model. This section uses the original
worlds, and its comparison, reading against inferring within the same
worlds, holds whichever factor is at work.

\textbf{Table 4.} Honest situations, source question, split by which
round the record checks. Gap is answer printed minus with record; the
last column is the extra loss in accuracy when the checked round changes
from 5 to 6, paired over worlds.

{\def\LTcaptype{none} 
\begingroup\small\begin{longtable}[]{@{}
  >{\raggedright\arraybackslash}p{(0.995\linewidth - 12\tabcolsep) * \real{0.0962}}
  >{\raggedright\arraybackslash}p{(0.995\linewidth - 12\tabcolsep) * \real{0.1923}}
  >{\raggedright\arraybackslash}p{(0.995\linewidth - 12\tabcolsep) * \real{0.0962}}
  >{\raggedright\arraybackslash}p{(0.995\linewidth - 12\tabcolsep) * \real{0.0962}}
  >{\raggedright\arraybackslash}p{(0.995\linewidth - 12\tabcolsep) * \real{0.1923}}
  >{\raggedright\arraybackslash}p{(0.995\linewidth - 12\tabcolsep) * \real{0.0962}}
  >{\raggedright\arraybackslash}p{(0.995\linewidth - 12\tabcolsep) * \real{0.2308}}@{}}
\toprule\noalign{}
\begin{minipage}[b]{\linewidth}\raggedright
model
\end{minipage} & \begin{minipage}[b]{\linewidth}\raggedright
situation
\end{minipage} & \begin{minipage}[b]{\linewidth}\raggedright
round
\end{minipage} & \begin{minipage}[b]{\linewidth}\raggedright
answer printed
\end{minipage} & \begin{minipage}[b]{\linewidth}\raggedright
with record
\end{minipage} & \begin{minipage}[b]{\linewidth}\raggedright
gap
\end{minipage} & \begin{minipage}[b]{\linewidth}\raggedright
gap change, r5 to r6
\end{minipage} \\
\midrule\noalign{}
\endhead
\bottomrule\noalign{}
\endlastfoot
Qwen 32B & nothing moved, honest & 5 & 1.000 & 0.953 & 0.047 & \\
& & 6 & 1.000 & 0.734 & 0.266 & +0.219 {[}+0.132, +0.306{]} \\
& points moved, honest & 5 & 0.969 & 0.891 & 0.078 & \\
& & 6 & 0.945 & 0.625 & 0.320 & +0.242 {[}+0.159, +0.320{]} \\
Qwen 72B & nothing moved, honest & 5 & 1.000 & 0.836 & 0.164 & \\
& & 6 & 1.000 & 0.398 & 0.602 & +0.438 {[}+0.286, +0.594{]} \\
& points moved, honest & 5 & 0.953 & 0.648 & 0.305 & \\
& & 6 & 0.820 & 0.188 & 0.633 & +0.328 {[}+0.115, +0.525{]} \\
Llama 70B & nothing moved, honest & 5 & 0.961 & 0.938 & 0.023 & \\
& & 6 & 0.961 & 0.547 & 0.414 & +0.391 {[}+0.273, +0.511{]} \\
& points moved, honest & 5 & 0.836 & 0.742 & 0.094 & \\
& & 6 & 0.648 & 0.305 & 0.344 & +0.250 {[}+0.116, +0.375{]} \\
\end{longtable}\endgroup
}

In all six rows the loss beyond reading widens from round 5 to round 6,
with every interval excluding zero. The answer-printed score moves too,
in the \emph{points moved} situation, and Llama's is low there (0.648 at
round 6); we report every with-record number beside its own
answer-printed number for that reason. But the widening gap is not a
reading effect. In \emph{nothing moved, honest}, the answer-printed
score is 0.96 to 1.00 at both rounds for all three models, and the
with-record score still falls by 0.22 to 0.44.

In the lying situations there is no round effect at all: 0.930 to 1.000
at both rounds in every model (Appendix C, Table C1).

\subsection{4.4 What moves the honest verdict, and what does
not}\label{what-moves-the-honest-verdict-and-what-does-not}

With round and letter crossed (Section 4.3), each factor gets its own
accuracy difference. Table 5 reports both.

\textbf{Table 5.} Honest situations, source question, with record,
original and reversed worlds together. Round effect is accuracy with the
record on round 5 minus accuracy with it on round 6. Letter effect is
accuracy when ``honest'' is the earlier pool letter minus when it is the
later. Paired over worlds.

{\def\LTcaptype{none} 
\begingroup\small\begin{longtable}[]{@{}
  >{\raggedright\arraybackslash}p{(0.995\linewidth - 6\tabcolsep) * \real{0.0694}}
  >{\raggedright\arraybackslash}p{(0.995\linewidth - 6\tabcolsep) * \real{0.2639}}
  >{\raggedright\arraybackslash}p{(0.995\linewidth - 6\tabcolsep) * \real{0.3194}}
  >{\raggedright\arraybackslash}p{(0.995\linewidth - 6\tabcolsep) * \real{0.3472}}@{}}
\toprule\noalign{}
\begin{minipage}[b]{\linewidth}\raggedright
model
\end{minipage} & \begin{minipage}[b]{\linewidth}\raggedright
situation
\end{minipage} & \begin{minipage}[b]{\linewidth}\raggedright
round effect (r5 minus r6)
\end{minipage} & \begin{minipage}[b]{\linewidth}\raggedright
letter effect (earlier minus later)
\end{minipage} \\
\midrule\noalign{}
\endhead
\bottomrule\noalign{}
\endlastfoot
Qwen 32B & nothing moved, honest & +0.238 {[}+0.176, +0.300{]} & $-$0.020
{[}-0.094, +0.058{]} \\
& points moved, honest & +0.277 {[}+0.215, +0.335{]} & $-$0.012 {[}-0.094,
+0.063{]} \\
Qwen 72B & nothing moved, honest & +0.348 {[}+0.244, +0.458{]} & +0.090
{[}-0.034, +0.216{]} \\
& points moved, honest & +0.328 {[}+0.203, +0.449{]} & +0.133 {[}-0.004,
+0.267{]} \\
Llama 70B & nothing moved, honest & +0.176 {[}+0.082, +0.267{]} & +0.215
{[}+0.127, +0.307{]} \\
& points moved, honest & +0.117 {[}-0.013, +0.237{]} & +0.320 {[}+0.200,
+0.435{]} \\
\end{longtable}\endgroup
}

The two families are moved by different things. For both Qwen models the
round is the lever. A record on round 6 costs 0.24 to 0.35, and the
letter margin is within noise at 32B and at most a small nudge at 72B
(+0.09 and +0.13, intervals reaching zero). For Llama the letter is the
lever: when ``honest'' is the later of the two pool letters, accuracy
drops by 0.21 to 0.32, and the round effect is smaller: +0.176 in
\emph{nothing moved, honest} and within noise in \emph{points moved,
honest}. We have no mechanism for either and offer none. What we can say
is that in both families the verdict on an honest reporter moves with a
feature that carries no information about the reporter, and in neither
family does the verdict on a lying reporter move by more than 0.035
(Appendix C, Table C1).

This is the shape of the finding. Detecting a lie is close to a fixed
computation: it survives the round, the letter, and the model family,
and at the 70B class it survives the wording too. Clearing an honest
reporter is not. It is a judgment the model reaches less often, and by a
route that irrelevant features can block.

Wording shows the same contrast. Within one model the honest situations
swing by 0.28 to 0.59 between the four wordings; the lying ones swing by
0.27 at 32B and by 0.02 or less at the 70B class (Appendix C, Table C2).

\subsection{4.5 A record that agrees with the reporter raises suspicion
anyway}\label{a-record-that-agrees-with-the-reporter-raises-suspicion-anyway}

The answer-printed-plus-record prompt differs from the answer-printed
prompt in one block: a rules line defining verified records, and the
record itself, are added to a prompt that already states the answer
(verified by diffing all 1,024 pairs). The two record lines differ in
one way. For the world of Section 2.1, the with-record version reads

\begin{verbatim}
Round 5 - actual result: 0 points. Source 7 had reported 0 points.
\end{verbatim}

and the answer-printed-plus-record version reads

\begin{verbatim}
Round 5 - actual result: 0 points.
\end{verbatim}

so in the answer-printed-plus-record prompt the model has to look up the
report in the log to see that the two agree. In the honest situations
they do agree. Adding a record whose value matches the reporter's should
not lower the score.

\textbf{Table 6.} Accuracy with the answer printed, and with the answer
printed plus the record, honest situations.

{\def\LTcaptype{none} 
\begingroup\small\begin{longtable}[]{@{}llll@{}}
\toprule\noalign{}
model & situation & answer printed & answer printed plus record \\
\midrule\noalign{}
\endhead
\bottomrule\noalign{}
\endlastfoot
Qwen 32B & nothing moved, honest & 1.000 & 0.996 \\
& points moved, honest & 0.957 & 0.949 \\
Qwen 72B & nothing moved, honest & 1.000 & 0.922 \\
& points moved, honest & 0.887 & 0.836 \\
Llama 70B & nothing moved, honest & 0.961 & 0.707 \\
& points moved, honest & 0.742 & 0.684 \\
\end{longtable}\endgroup
}

Paired over worlds, the difference, answer printed minus answer printed
plus record, in \emph{nothing moved, honest} is +0.004 {[}+0.000,
+0.012{]} at Qwen 32B, +0.078 {[}+0.043, +0.121{]} at Qwen 72B, and
+0.254 {[}+0.176, +0.340{]} at Llama 70B. Told in so many words that the
source did not begin lying, and then shown a record whose value matches
what the source reported, Llama says the source is lying 29\% of the
time. In the lying situations the difference is within 0.012 of zero for
every model. So the record's presence changes nothing at 32B, costs a
little at 72B, and costs Llama a quarter of its honest verdicts, and in
no model does it move the lying verdict. We report this as a per-model
effect and not as a property of language models.

\subsection{4.6 Saying the two causes are equally common changes
nothing}\label{saying-the-two-causes-are-equally-common-changes-nothing}

The salience version adds one rules line, which says that across
histories of this kind the values swapped about as often as a source
began lying. If the models held a blame-the-messenger prior that
evidence has to overcome, making the alternative salient should help.

It did not. In the \emph{points moved, honest} situation the change from
with record to salience is $-$0.086 {[}-0.129, $-$0.047{]} at Qwen 32B,
+0.000 {[}-0.031, +0.027{]} at Qwen 72B, and +0.027 {[}+0.004, +0.051{]}
at Llama 70B. Across all four situations it is $-$0.034, $-$0.010, and
$-$0.017. We registered +0.10 as the threshold that would favor a movable
prior and +0.05 as the threshold below which the world-change story is
simply not being entertained. All three land below +0.05.

One reading is that the alternative is not weighed and lost but never
entertained. A null against a small registered threshold is weak
evidence for that, so we hold the reading loosely. The line also names
both causes, and a reminder that mentions ``began lying'' may itself cue
the lying story.

\subsection{4.7 The 8B model cannot read the
task}\label{the-8b-model-cannot-read-the-task}

Llama-3.1-8B scores 0.620 with the answer printed on the source question
and 0.778 on the fresh-pair question, both below the registered 0.90
bar. By rule 2, none of its other numbers are read.

\subsection{4.8 Knowing is not acting}\label{knowing-is-not-acting}

Every history is also asked which of the fresh pair pays. In the lying
situations the correct answer is the option the source reported as
losing. The two questions are asked in two separate prompts that share
every byte before the question line. So for each history we can pair the
answers and ask: on histories where the model's answer to the honesty
question was correct, does its answer to the fresh-pair question act on
that verdict?

\textbf{Table 7.} Fresh-pair accuracy on the histories where the same
model, in the separate honesty prompt, correctly said the source is
lying. With record, byte-matched pairs. The last two columns are a
different statistic: how often the fresh-pair answer is the option the
source reported as winning, in the same two situations.

{\def\LTcaptype{none} 
\begingroup\small\begin{longtable}[]{@{}
  >{\raggedright\arraybackslash}p{(0.995\linewidth - 8\tabcolsep) * \real{0.1111}}
  >{\raggedright\arraybackslash}p{(0.995\linewidth - 8\tabcolsep) * \real{0.2778}}
  >{\raggedright\arraybackslash}p{(0.995\linewidth - 8\tabcolsep) * \real{0.2778}}
  >{\raggedright\arraybackslash}p{(0.995\linewidth - 8\tabcolsep) * \real{0.1667}}
  >{\raggedright\arraybackslash}p{(0.995\linewidth - 8\tabcolsep) * \real{0.1667}}@{}}
\toprule\noalign{}
\begin{minipage}[b]{\linewidth}\raggedright
model
\end{minipage} & \begin{minipage}[b]{\linewidth}\raggedright
nothing moved, lying
\end{minipage} & \begin{minipage}[b]{\linewidth}\raggedright
points moved, lying
\end{minipage} & \begin{minipage}[b]{\linewidth}\raggedright
follows the report, nothing moved
\end{minipage} & \begin{minipage}[b]{\linewidth}\raggedright
follows the report, points moved
\end{minipage} \\
\midrule\noalign{}
\endhead
\bottomrule\noalign{}
\endlastfoot
Qwen 32B & 0.117 {[}0.085, 0.150{]} & 0.200 {[}0.148, 0.255{]} & 0.887 &
0.812 \\
Qwen 72B & 0.145 {[}0.102, 0.191{]} & 0.180 {[}0.129, 0.230{]} & 0.855 &
0.820 \\
Llama 70B & 0.369 {[}0.311, 0.427{]} & 0.439 {[}0.373, 0.502{]} & 0.629
& 0.562 \\
\end{longtable}\endgroup
}

On the very histories where the Qwen models correctly said that the
source reports the opposite of every result, asked separately which
fresh option pays, they take the source's report at face value more than
80\% of the time. The answer-printed score on the second question is
0.987 to 0.994, so this is not a failure to read the options. Llama acts
on its own diagnosis more often, and still below half.

This is the knowing-doing gap that Schmied et al.~{[}2025{]} report in
bandit tasks, where models state the right rationale and take the greedy
action anyway, and we reproduce its shape rather than claim it. What our
version adds is the setting. The knowledge here is a verdict about the
reward channel, and acting on it means reversing what the channel
reported for the fresh pair, in the same history.

\section{5. What was registered, and what
happened}\label{what-was-registered-and-what-happened}

We fixed the analysis before the runs and wrote down predictions. Three
of them deserve a plain account.

The central prediction, the with-record score on \emph{points moved,
honest} at Qwen 72B, was registered at 0.65. We had a 16-world
measurement of 0.422 on \emph{points moved, honest} at Qwen 72B, and
three other 16-world numbers had shrunk or vanished at 64 worlds, so we
predicted this one would too. The rule we fixed said that below 0.55,
with an interval excluding 0.60, the deficit is real. The 64-world
result was 0.418 {[}0.316, 0.523{]}. The prediction was wrong; the
larger run gave nearly the same score as the small one; the rule fired.

The round split was a pre-registered estimator, but the emphasis on it
in Section 4.3 is ours after the fact, and Section 4.4 exists because
the estimator was confounded as built. We report the confound, the fix,
and both sets of worlds.

The third account is the two bars, rule by rule.

Rule 1 was registered as ``a calibrated arm outscoring its own ceiling
voids the run.'' The calibrated arm is the with-record version and the
ceiling is answer printed plus record. Applied to each model's whole
row, no model fails: with record is below its ceiling at every model on
both questions. Applied per situation, two Llama cells fail:
\emph{nothing moved, honest} (0.742 against 0.707) and \emph{points
moved, lying} (0.996 against 0.992, a difference of one prompt in 256).

Rule 2 was registered as ``the honest reading floor below 0.90 voids the
winner arm.'' The floor is the answer-printed score and the winner arm
is the fresh-pair question. A later amendment for the 72B raised its bar
to 0.95 and said a failure makes ``the row uninterpretable.'' Llama ran
after both and had no registered bar of its own. We have applied the
0.90 bar to the source question as well, which is stricter than the
registered wording. At the row level every large model clears it: the
answer-printed source scores are 0.987 at Qwen 32B, 0.972 at Qwen 72B,
and 0.921 at Llama 70B. Applied per situation, two cells fail, both
\emph{points moved, honest}: 0.887 at Qwen 72B and 0.742 at Llama 70B.

Neither rule said whether it applied to a row or to each situation, and
we first read them the way that flattered the result. Read per
situation, four cells are discounted, and here is what survives. At Qwen
32B every cell passes, so the penalty stands under either reading. At
Qwen 72B \emph{nothing moved, honest} passes with an answer-printed
score of 1.000, so the penalty stands on that situation alone (1.000
against 0.617). At Llama 70B both honest cells fail a per-situation
reading, one on each rule, so the Llama penalty rests on the row-level
reading and on nothing else. A reader who applies the rules per
situation has two models, not three. That is why every number in the
paper is printed beside the answer-printed score of its own situation.

\section{6. Related work}\label{related-work}

\textbf{The setting.} Everitt et al.~{[}2017{]} define the corrupt
reward Markov decision process, a game in which the reward the agent
sees may differ from the true one, show that no agent can do well in it
in general (their Theorem 11), and identify richer data about the reward
channel as the way out. Bandit work on adversarial corruption
{[}Lykouris et al., 2018; and successors{]} bounds regret under
corrupted rewards without asking the agent to diagnose the corruption.
We are not aware of prior work that places a language model inside this
setting and hands it the richer data as a single verified record.

\textbf{Knowing without acting.} Schmied et al.~{[}2025{]} name the
knowing-doing gap and measure it in bandit tasks with the same instances
and only the instruction differing, finding that models state the
correct rationale and act greedily anyway. Section 4.8 reproduces that
shape. Related dissociations between what a model can state and what it
does have been reported for parametric knowledge {[}Cultural Binding
Heads, 2026{]} and for premise correction {[}Knowing but Not Correcting,
2026{]}.

\textbf{Surface sensitivity.} That language models' answers move with
the letters and positions of multiple-choice options is well documented
{[}Zheng et al., 2024{]}, and recent work argues that primacy and
anchoring effects are structural in autoregressive models {[}Bias by
Necessity, 2026; Anchors in the Machine, 2025{]}. Our contribution is
not that these effects exist but that they act on one of two symmetric
judgments and not the other.

\textbf{Inverse scaling.} McKenzie et al.~{[}2023{]} catalog tasks on
which larger models do worse because a strong prior overrides in-context
evidence; the ordering in Table 2 has that shape, and Section 4.2 says
why we make no claim of it.

\section{7. Limits}\label{limits}

Three models from two families clear the reading bar. That is enough to
say the honest-reporter penalty is not one family's quirk and not enough
to say anything about size. Every number is from one seed of 64 worlds.

The four wordings are ours, and the letter and position effects show
that wording matters; a fifth wording could move the honest numbers
again, though at the 70B class nothing we tried moved the lying ones.

The one-token protocol measures a single-step read, and a model allowed
to reason aloud might clear an honest reporter more often; that is a
different question, and the single-step read is the one a policy makes.
The salience result has a confound we name in Section 4.6. And we have
no mechanism for why the round moves Qwen and the letter moves Llama; we
report the margins and stop.

\section{8. What it means}\label{what-it-means}

Everitt's remedy for a corrupted reward channel is richer data about the
channel. The implicit assumption is that the agent uses that data the
way a statistician would, to raise its estimate of the channel's honesty
when the check passes and lower it when the check fails. The three large
models we tested do the second reliably and the first unreliably, and
how unreliably depends on which round was checked or which letter stands
for ``honest.'' That is what we observed. The consequence is for an
agent that acts on the verdict: a verification step in front of it will
catch a broken reporter, and will also, some fraction of the time that
depends on nothing relevant, convict an honest one, after which the
agent would invert a channel that was telling it the truth. We did not
observe that last step; Section 4.8 shows the models tested here mostly
do not act on their verdicts in either direction.

\begin{sloppypar}
Code, prompts, answer keys, and every scored output are released at
\mbox{\url{https://github.com/IamArmanNikkhah/easy-to-catch-a-liar}}.
\end{sloppypar}

\section{References}\label{references}

Everitt, T., Krakovna, V., Orseau, L., Hutter, M., and Legg, S. (2017).
Reinforcement learning with a corrupted reward channel. \emph{IJCAI
2017}. arXiv:1705.08417.

Grattafiori, A. et al.~(2024). The Llama 3 herd of models.
arXiv:2407.21783.

Lykouris, T., Mirrokni, V., and Paes Leme, R. (2018). Stochastic bandits
robust to adversarial corruptions. \emph{STOC 2018}. arXiv:1803.09353.

McKenzie, I. et al.~(2023). Inverse scaling: When bigger isn't better.
arXiv:2306.09479.

Qwen Team (2025). Qwen2.5 technical report. arXiv:2412.15115.

Schmied, T., Bornschein, J., Grau-Moya, J., Wulfmeier, M., and Pascanu,
R. (2025). LLMs are greedy agents: Effects of RL fine-tuning on
decision-making abilities. arXiv:2504.16078.

Zheng, C., Zhou, H., Meng, F., Zhou, J., and Huang, M. (2024). Large
language models are not robust multiple choice selectors. \emph{ICLR
2024}. arXiv:2309.03882.

Floro, A. and Benedetto, L. (2026). Cultural binding heads in language
models. arXiv:2605.28543.

Chen, Z., Lin, H., Chen, Z., Tian, Y., Yang, G., Wang, D., Guo, Y., Zhu,
H., and Cheng, J. (2026). Knowing but not correcting: Routine task
requests suppress factual correction in LLMs. arXiv:2605.05957.

Wu, J., Guo, D., and Yiu, S.-M. (2026). Bias by necessity: Impossibility
theorems for sequential processing with convergent AI and human
validation. arXiv:2605.08716.

Valencia-Clavijo, F. (2025). Anchors in the machine: Behavioral and
attributional evidence of anchoring bias in LLMs. arXiv:2511.05766.

\section{Appendix A. One full prompt}\label{appendix-a.-one-full-prompt}

The with-record source prompt for one \emph{points moved, honest} world
in the log wording, exactly as scored:

\begin{verbatim}
You are choosing between labelled options to earn points.

Rules:
- On each round you pick one of the two options available that round. Exactly one of them is worth 1 point and the other is worth 0 points.
- A reporting source tells you the result of each round, either 1 point or 0 points.
- Through round 4 every source reported each result accurately, and the point values did not change.
- After round 4, any of these may have happened: the point values of the two original options swapped; a source started reporting the opposite of each actual result; both; or neither. You are not told which.
- If a source started reporting the opposite, it keeps doing so on every later round.
- A new pair of options may be introduced later. A new pair has its own point values, again exactly one worth 1 point.
- A verified record states the actual result of a round, independently of what a source said about that round.

Verified records:
Round 5 - actual result: 0 points. Source 7 had reported 0 points.

Round log:
Round 1 - you chose M. Source 7 reported 1 point.
Round 2 - you chose W. Source 7 reported 0 points.
Round 3 - you chose M. Source 7 reported 1 point.
Round 4 - you chose W. Source 7 reported 0 points.
Round 5 - you chose M. Source 7 reported 0 points.
Round 6 - you chose W. Source 7 reported 1 point.
Round 7 - you chose C. Source 7 reported 1 point.
Round 8 - you chose D. Source 7 reported 0 points.

Question: Since round 5, has Source 7 been reporting each result accurately, or reporting the opposite of each actual result?
Reply K if accurately, Q if the opposite.
Reply with a single letter, K or Q, and nothing else.
Answer:
\end{verbatim}

\section{Appendix B. Every number}\label{appendix-b.-every-number}

\texttt{paper/NUMBERS.md} is generated by
\texttt{code/paper\_numbers.py} from the scored outputs and is the only
source for every figure in this paper.

\section{Appendix C. Two tables that confirm the lying verdict does not
move}\label{appendix-c.-two-tables-that-confirm-the-lying-verdict-does-not-move}

\textbf{Table C1.} Lying situations, round and letter crossed as in
Table 5. Every cell is between 0.930 and 1.000 and every margin is
within 0.035 of zero. The Qwen 72B rows are exactly 1.000 in every cell,
so their margins have no spread.

{\def\LTcaptype{none} 
\begingroup\small\begin{longtable}[]{@{}
  >{\raggedright\arraybackslash}p{(0.995\linewidth - 6\tabcolsep) * \real{0.0806}}
  >{\raggedright\arraybackslash}p{(0.995\linewidth - 6\tabcolsep) * \real{0.2903}}
  >{\raggedright\arraybackslash}p{(0.995\linewidth - 6\tabcolsep) * \real{0.3065}}
  >{\raggedright\arraybackslash}p{(0.995\linewidth - 6\tabcolsep) * \real{0.3226}}@{}}
\toprule\noalign{}
\begin{minipage}[b]{\linewidth}\raggedright
model
\end{minipage} & \begin{minipage}[b]{\linewidth}\raggedright
situation
\end{minipage} & \begin{minipage}[b]{\linewidth}\raggedright
round effect (r5 minus r6)
\end{minipage} & \begin{minipage}[b]{\linewidth}\raggedright
letter effect (earlier minus later)
\end{minipage} \\
\midrule\noalign{}
\endhead
\bottomrule\noalign{}
\endlastfoot
Qwen 32B & nothing moved, lying & $-$0.004 {[}-0.041, +0.034{]} & $-$0.004
{[}-0.042, +0.034{]} \\
& points moved, lying & +0.000 {[}-0.037, +0.038{]} & +0.000 {[}-0.038,
+0.039{]} \\
Qwen 72B & nothing moved, lying & +0.000 & +0.000 \\
& points moved, lying & +0.000 & +0.000 \\
Llama 70B & nothing moved, lying & +0.027 {[}+0.008, +0.051{]} & $-$0.020
{[}-0.043, $-$0.000{]} \\
& points moved, lying & +0.023 {[}+0.003, +0.048{]} & $-$0.031 {[}-0.057,
$-$0.011{]} \\
\end{longtable}\endgroup
}

\textbf{Table C2.} Source accuracy with the record, by wording, original
worlds. The honest situations move with the wording; the lying ones
barely do at the 70B class.

{\def\LTcaptype{none} 
\begingroup\small\begin{longtable}[]{@{}llllll@{}}
\toprule\noalign{}
model & situation & commentary & log & minutes & prose \\
\midrule\noalign{}
\endhead
\bottomrule\noalign{}
\endlastfoot
Qwen 32B & nothing moved, honest & 0.891 & 0.797 & 0.688 & 1.000 \\
& points moved, honest & 0.875 & 0.750 & 0.406 & 1.000 \\
& nothing moved, lying & 1.000 & 1.000 & 1.000 & 0.734 \\
& points moved, lying & 1.000 & 1.000 & 1.000 & 0.750 \\
Qwen 72B & nothing moved, honest & 0.812 & 0.672 & 0.375 & 0.609 \\
& points moved, honest & 0.531 & 0.484 & 0.203 & 0.453 \\
& nothing moved, lying & 1.000 & 1.000 & 1.000 & 1.000 \\
& points moved, lying & 1.000 & 1.000 & 1.000 & 1.000 \\
Llama 70B & nothing moved, honest & 0.688 & 0.828 & 0.641 & 0.812 \\
& points moved, honest & 0.375 & 0.641 & 0.422 & 0.656 \\
& nothing moved, lying & 1.000 & 1.000 & 1.000 & 0.984 \\
& points moved, lying & 1.000 & 1.000 & 1.000 & 0.984 \\
\end{longtable}\endgroup
}

\end{document}